\documentclass[conference]{IEEEtran}
\usepackage{times}
\usepackage{caption}
\usepackage{afterpage}
\usepackage{float}
\usepackage{subcaption}
\usepackage{placeins} 

\usepackage{graphicx,float,booktabs,xcolor,multirow,array,color,ifthen,tabu,colortbl,dblfloatfix,url,xparse,mathtools,patchcmd,algorithm,algorithmic,amssymb,xspace,nicefrac,microtype,amsmath,amsthm,amsfonts,bm,ragged2e,tikz,stackengine,etoolbox,xpatch,enumerate,xstring,setspace,makecell,changepage,cuted,titlesec, wrapfig,tcolorbox, wrapfig}
\usepackage[numbers]{natbib}
\usepackage{multicol, booktabs}
\usepackage{afterpage}

\usepackage[pagebackref=true,breaklinks=true,colorlinks=true,bookmarks=false,citecolor=orange,linkcolor=orange,urlcolor=teal]{hyperref}

\author{\vspace{2mm}
    Tony Tao\IEEEauthorrefmark{1}, Mohan Kumar Srirama\IEEEauthorrefmark{1}, Jason Jingzhou Liu, Kenneth Shaw, Deepak Pathak \\\vspace{.5mm}
    Carnegie Mellon University \\
    \IEEEauthorrefmark{1}Equal contribution
}

\newcommand{\our}{DexWild\xspace}
\newcommand{\ours}{DexWilds\xspace}
\newcommand{\oursystem}{DexWild-System\xspace}

\begin{document}

\title{\textbf{DexWild}: Dexterous Human Interactions for In-the-Wild Robot Policies}

\makeatletter
\let\@oldmaketitle\@maketitle
\renewcommand{\@maketitle}{\@oldmaketitle
  \includegraphics[width=1\linewidth]{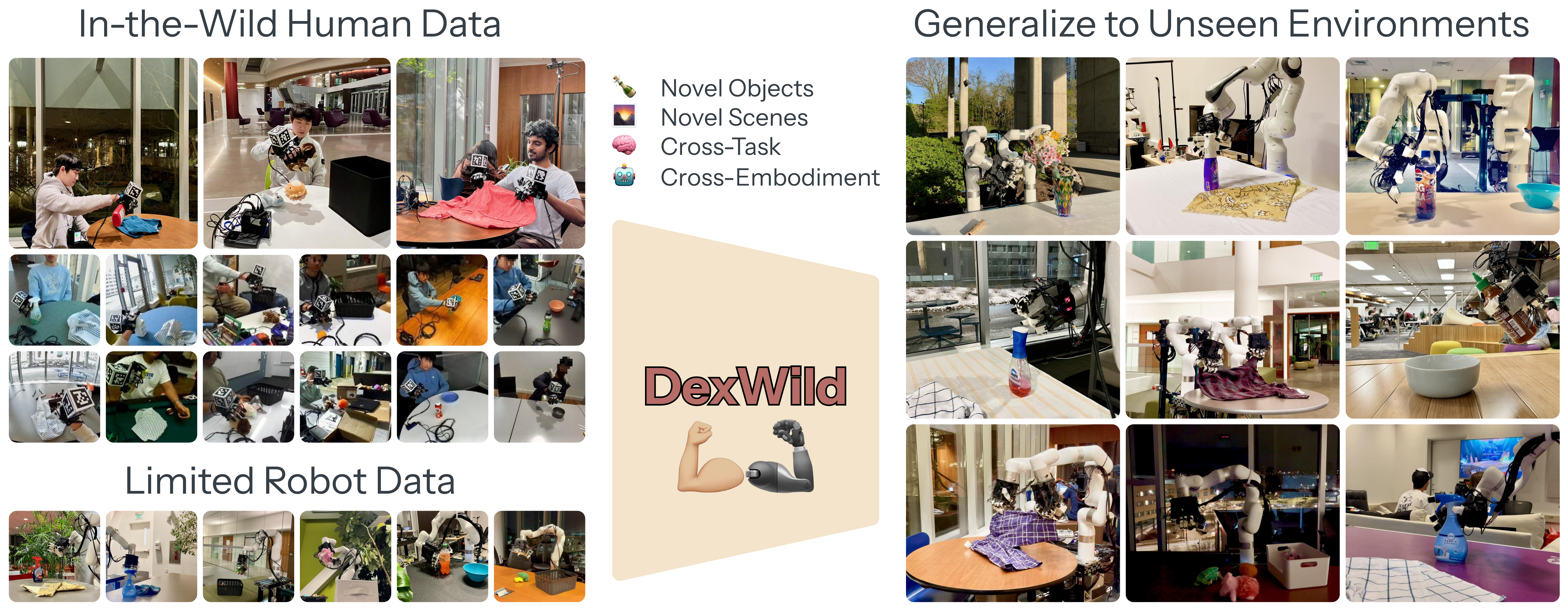}
  \centering
  \captionof{figure}{\small \textbf{\our} enables dexterous policies to generalize to new objects, scenes, and embodiments. This is achieved by leveraging large-scale, real-world human embodiment data collected in many scenes and co-trained with a smaller robot embodiment dataset for grounding.}
  \label{fig:teaser}
  }
\makeatother
\maketitle
\setcounter{figure}{1}

\begin{abstract}
Large-scale, diverse robot datasets have emerged as a promising path toward enabling dexterous manipulation policies to generalize to novel environments, but acquiring such datasets presents many challenges. While teleoperation provides high-fidelity datasets, its high cost limits its scalability.  Instead, what if people could use their own hands, just as they do in everyday life, to collect data?  In \our, a diverse team of data collectors uses their hands to collect hours of interactions across a multitude of environments and objects. To record this data, we create \oursystem, a low-cost, mobile, and easy-to-use device.  The \our learning framework co-trains on both human and robot demonstrations, leading to improved performance compared to training on each dataset individually.  This combination results in robust robot policies capable of generalizing to novel environments, tasks, and embodiments with minimal additional robot-specific data. Experimental results demonstrate that \our significantly improves performance, achieving a 68.5\% success rate in unseen environments—nearly four times higher than policies trained with robot data only—and offering 5.8$\times$ better cross-embodiment generalization.  Video results, codebases, and instructions at \url{https://dexwild.github.io}

\end{abstract}

\IEEEpeerreviewmaketitle

\section{Introduction}
Roboticists have long dreamed of creating robots that can perform tasks with the same dexterity and adaptability as humans. We would like robots to deftly generalize to many different objects, environments, and embodiments-yet this vision of truly versatile robot behaviors remains a formidable challenge. While there have been many breakthroughs in large language models (LLMs) ~\cite{vaswani2017attention, touvron2023llama, brown2020language} and vision language models (VLMs) ~\cite{liu2023llava, steiner2024paligemma2}, the key to their success lies in harnessing vast datasets. In contrast, robotics faces a critical hurdle: large-scale, diverse robot datasets needed to train foundation models do not yet exist.

In recent years, a key approach to collecting robot datasets has been through teleoperation, which provides high-precision, high-quality action data that a policy can directly train on. ~\cite{open_x_embodiment_rt_x_2023, khazatsky2024droid, walke2023bridgedata}. However, acquiring this data requires highly-trained human operators working with specialized robot setups. Gathering data in diverse environments presents additional challenges such as physically relocating the robot to each new location. This data collection process is both labor-intensive and expensive, making it difficult to scale to the volume of data needed for dexterous generalization in unseen environments.

Another approach to scaling robot datasets is to leverage internet-scale video data from platforms like YouTube, which provide vast and diverse visual grounding in real-world environments ~\cite{grauman2022ego4d, damen2018epic}. However, utilizing this data effectively presents significant challenges. First, publicly available videos often lack the fine-grained accuracy needed to capture detailed hand states because vision-based body detection modules are noisy and unreliable. Additionally, these videos are not inherently structured with categorized episodes for task-specific learning, further complicating their direct application in robotics. \cite{hu2023toward, bahl2023affordances, shaw2023videodex}.   While some data collection efforts exist with more accurate and structured data, \cite{zimmermann2019freihand, banerjee2024hot3d}, they do not have enough environment diversity.  We seek to collect data with \textbf{tracking accuracy and environment diversity} to enable generalizable dexterous behavior.

To overcome these barriers, some have explored collecting accurate in-the-wild human demonstrations by equipping users with a wearable gripper that directly maps their hand movements to robot actions~\cite{chi2024universal}. However, this approach is cumbersome, ill-suited for natural, everyday interactions, and constrains the collected data to a specific embodiment. Other works \cite{wang2024dexcap} propose using dexterous hands and gloves, but they do not scale to in-the-wild environments.

In this paper, we present \our, a system that enables effective learning of robust dexterous manipulation policies through co-training on human and robot demonstrations. Our key contributions include:
\begin{enumerate}
    \item \textbf{Scalable Data Collection System}: A novel human-embodiment \oursystem that enables untrained operators to quickly collect 9,290 demonstrations across 93 diverse environments, achieving 4.6$\times$ speedup over conventional robot-based methods
    \item \textbf{Efficient Co-training Framework}: An approach that optimally combines human and robot demonstrations, significantly improving policy generalization to achieve 68.5\% success rate in novel environments, nearly four times higher than robot-only policies.
    \item \textbf{Strong Cross Embodiment and Cross Task Performance}: Our data collection system combined with our co-training framework achieves of 5.8$\times$ improvement in cross-embodiment transfer over baselines and effective skill transfer across tasks.
\end{enumerate}

\section{Related Works} 

\begin{figure*}[t]
\centering
\includegraphics[width=\linewidth]{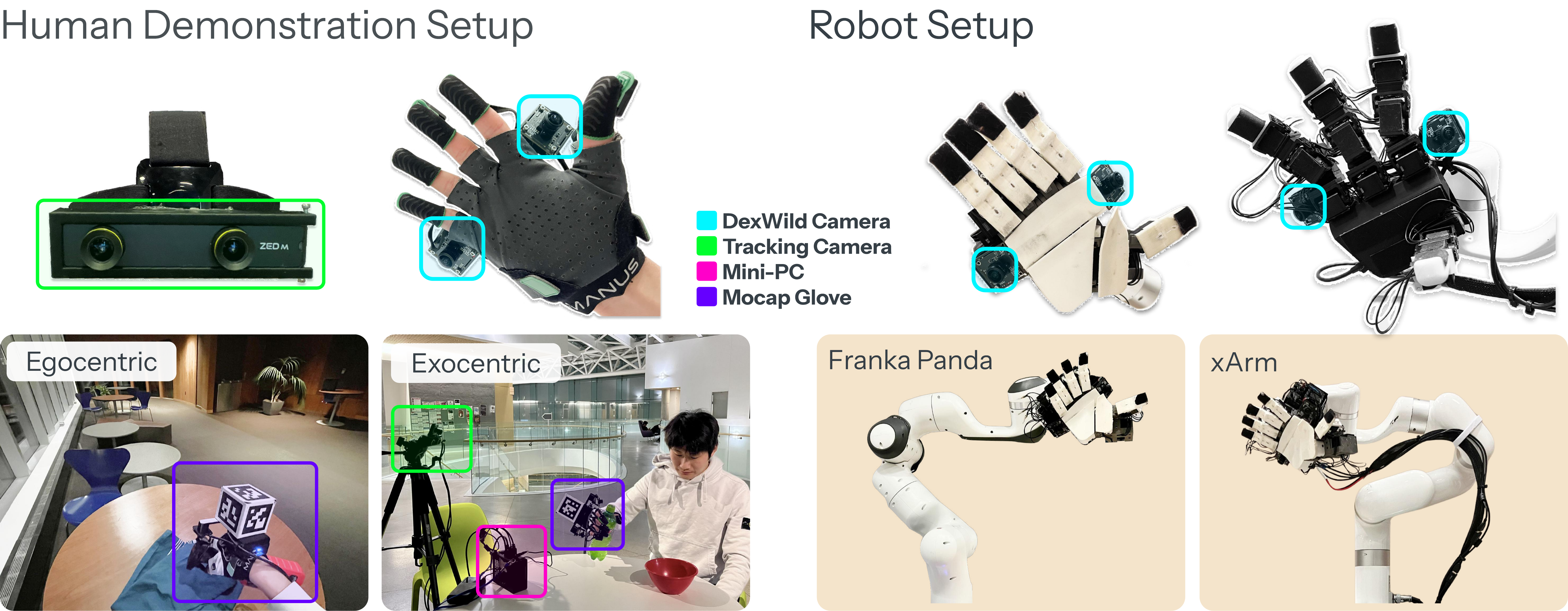}
\vspace{-0.2in}
\caption{\small \textbf{Left:} \our efficiently capture high-fidelity data using an individual’s own hands across various environments. \textbf{Right:} Robot hands are equipped with cameras aligned with the human cameras. We test \our on two distinct robot hands and robot arms.}
 \label{fig:hardware}
 \vspace{-0.15in}
\end{figure*}

\subsection{Generalization for Imitation Learning}
Learning generalizable policies for robot manipulation has seen rapid progress, driven largely by advances in visual representation learning and imitation learning from large-scale datasets. On the visual side, embodied representation learning has benefited from egocentric datasets such as Ego4D~\cite{grauman2022ego4d} and EPIC-KITCHENS~\cite{damen2018epic}, with recent methods~\cite{nair2022r3m, dasari2023datasets, srirama2024hrp, pmlr-v205-shaw23a} leveraging these datasets to train scalable visual encoders. However, these approaches still require substantial downstream robot demonstrations to train control policies.

In parallel, robot-only demonstration datasets have grown significantly in scale and diversity~\cite{khazatsky2024droid, open_x_embodiment_rt_x_2023, walke2023bridgedata}, fueling research in behavior cloning and enabling generalist policy architectures~\cite{om2023octo, open_x_embodiment_rt_x_2023, kim2024openvla}. While these policies show impressive performance across many tasks, they often struggle to generalize to unseen object categories, scene layouts, or environmental conditions~\cite{mandlekar2021matters}. This lack of robustness remains a key limitation of current systems.

\subsection{Data Generation for Robot Manipulation}
Overcoming the robot data bottleneck has become a central challenge in robot learning.

One approach leverages internet videos to extract action information. Several works, such as VideoDex \cite{shaw2023videodex} and HOP \cite{singh2024handobjectinteractionpretrainingvideos}, utilize large scale human videos to learn an action prior through retargeting, which they use to bootstrap policy training. Others, such as LAPA \cite{ye2024latent}, use unlabelled videos to generate latent action representations that can be used for downstream tasks. While these video‐based schemes enjoy vast visual diversity, they typically fall short at capturing the precise, low‐level motor commands needed for real‐world manipulation.

Simulation enables rapid generation of action data at scale. However, creating diverse, realistic environments for many tasks and addressing the sim-to-real gap is challenging. Recent successes in transferring manipulation policies from simulation \cite{singh2024dextrah} have been confined to tabletop settings and lack the generalization needed for deployment in diverse environments.

Direct teleoperation on physical robots yields the highest ﬁdelity, but scales poorly. Recent works have shown impressive dexterity and efficient learning in fixed scenarios \cite{zhao2023learning,wu2023gello,shaw2024bimanual, iyer2024open}, yet collecting enough demonstrations to generalize across diverse scenes quickly becomes prohibitively expensive.

Recently, there has been a growing body of work that utilizes purpose-collected high quality human embodiment data without the tedious teleoperation. We discuss these approaches in the next section.

\subsection{Human Action Tracking Systems}
In order to acquire high-quality data from human motions, accurate hand and wrist tracking is of paramount importance. To bypass the complexities of hand pose estimation, several works equip users with handheld robot grippers~\cite{chi2024universal, etukuru2024robotutilitymodelsgeneral, song2020grasping}. While this approach simplifies retargeting, it constrains users to the specific morphology of the robot gripper, limiting the diversity of captured behavior. Moreover, many of these systems rely on SLAM-based wrist tracking, which can fail in feature-sparse environments or when occlusions occur \cite{chi2024universal, lin2024datascaling}—such as during drawer opening or tool use.

Other approaches aim to estimate both hand and wrist poses directly from visual input~\cite{pavlakos2024reconstructing, rong2021frankmocap, cheng2024open, telekinesis, pavlakoshamer, kareer2024egomimic, qiu2025-humanpolicy}. These methods are easy to deploy and require no instrumentation, but their performance degrades significantly under occlusion—an unavoidable situation in manipulation. Alternative strategies for wrist tracking, such as IMU-based~\cite{corrales2008hybrid, tian2015upper} and outside-in optical systems~\cite{pfister2014comparative}, come with their own limitations: IMUs are lightweight and portable but prone to drift, while optical systems are accurate yet require laborious calibration and controlled environments. \our leverages calibration-free Aruco tracking—significantly improving reliability and minimizing setup time as it requires a single monocular camera.

While vision-based methods often attempt to track both the wrist and fingers simultaneously, many recent systems decouple the two to improve accuracy. Kinematic exoskeleton gloves can provide high-fidelity joint measurements and even haptic feedback~\cite{zhang2025doglove}, but are bulky and uncomfortable for long-term use. Instead, \our, along with prior works~\cite{shaw2024bimanual, wang2024dexcap}, adopts a lightweight glove-based solution that uses electromagnetic field (EMF) sensing to estimate fingertip positions. This allows for accurate, real-time hand tracking that is robust to occlusions and readily retargetable to a wide range of robot hands.

\section{\our}

Many believe that leveraging large, high-quality datasets is the key for creating dexterous robot policies that generalize ~\cite{open_x_embodiment_rt_x_2023, om2023octo, shaw2023videodex, dasari2023datasets}. We introduce \oursystem, a user-friendly, high-fidelity platform for efficiently gathering natural human hand demonstrations across diverse real-world settings. Compared to traditional teleoperation-based approaches, \oursystem enables 4.6$\times$ faster data acquisition at scale.

Building on this system, we propose \our, an imitation learning framework that co-trains on large-scale \oursystem human demonstrations alongside a small number of robot demonstrations. This approach combines the diversity and richness of human interactions with the grounding of the robot embodiment, enabling policies to robustly generalize across new objects, environments, and embodiments. Figure \ref{fig:teaser} displays our high level approach.

\subsection{Data Collection System}

A scalable data collection system for dexterous robot learning must enable natural, efficient, and high-fidelity collection across diverse environments. To this end, we design \oursystem: a portable, user-friendly system that captures human dexterous behavior with minimal setup and training. While previous in-the-wild data collection approaches have typically relied on sensorized grippers, we aimed to create a more intuitive hardware interface that mirrors how humans naturally interact with the world. From delicate fine-motor actions to powerful grasps, humans possess dexterity across a wide range of manipulation tasks. By learning from this intrinsic capability, \oursystem captures rich, diverse data applicable to a broad range of robot embodiments.

\oursystem is designed around three core objectives: 
\begin{itemize} 
\item \textbf{Portability:} Allow rapid, large-scale data collection across diverse environments without requiring complex calibration procedures. 
\item \textbf{High Fidelity:} Accurately capture fine-grained hand and environment interactions essential for training precise dexterous policies. 
\item \textbf{Embodiment-Agnostic:} Enable seamless retargeting from human demonstrations to a wide variety of robot hands. 
\end{itemize}

\textbf{Portability:}

To collect data in diverse real-world settings, a system must be portable, robust, and usable by anyone. We design \oursystem with these goals in mind: it is lightweight, easy to carry, and can be set up in just a few minutes—enabling scalable data collection across many locations. 

As shown in Figure~\ref{fig:hardware}, \oursystem consists of only three components: a single tracking camera for wrist pose estimation, a battery-powered mini-PC for onboard data capture, and a custom sensor pod comprising a motion-capture glove and synchronized palm-mounted cameras.

Unlike traditional motion capture systems \cite{zimmermann2019freihand, fan2023arctic, chao2021dexycb, steamvr} that often rely on complex outside-in tracking setups that require calibration, \oursystem is truly calibration free, making it versatile for any scenario and foolproof for untrained operators.

This is achieved by adopting a relative state-action representation, where each state and action is captured as the relative difference from the previous time step's pose. This eliminates any need for a global coordinate frame, allowing the tracking camera to be freely placed—either egocentrically or exocentrically. Additionally, the palm cameras are rigidly mounted in fixed positions across both human and robot embodiments. This ensures visual observations are aligned across domains, eliminating the need for further calibration at deployment. The external tracking camera, when carefully positioned, can also capture supplementary environmental context useful for learning robust policies.

\textbf{High Fidelity:}

To learn dexterous behaviors, fine-grained, nuanced motions must be captured in the training dataset. Although \oursystem consists of only a few portable components, we make no compromises on data fidelity. Our system is designed to accurately capture both hand and wrist actions, paired with high-quality visual observations.

For wrist and hand tracking, vision-only methods are easy to setup. However, what they gain in portability, they often lose in accuracy and robustness—yielding noisy pose estimates that degrade policy learning~\cite{shaw2024bimanual, open2024television, qiu2025-humanpolicy, chi2024universal}. 

For hand pose estimation, we use motion capture gloves, which offer high accuracy, low latency, and robustness against occlusions~\cite{shaw2024bimanual}. For wrist tracking, we mount ArUco markers on the glove and track them using an external camera. This avoids the fragility of SLAM-based wrist tracking, which often fails in feature-sparse environments or during occlusion-heavy tasks (e.g., drawer opening).

Unlike many datasets that rely on egocentric or distant external cameras, we place two global-shutter cameras directly on the palm. As illustrated in Figure~\ref{fig:hardware}, these stereo cameras capture detailed, localized interaction views with minimal motion blur and a wide field of view. This wide field of view enables policies to operate using only the onboard palm cameras, without any reliance on static viewpoints.

\textbf{Embodiment-Agnostic:} 

To ensure the longevity and versatility of \our data, we aim for it to remain useful across different robot embodiments—even as hardware platforms evolve. Achieving this goal requires careful alignment of both the observation space and the action space between humans and robots.

\begin{figure}[t]
\begin{minipage}[b]{0.5\textwidth} 
\centering
\includegraphics[width=\linewidth]{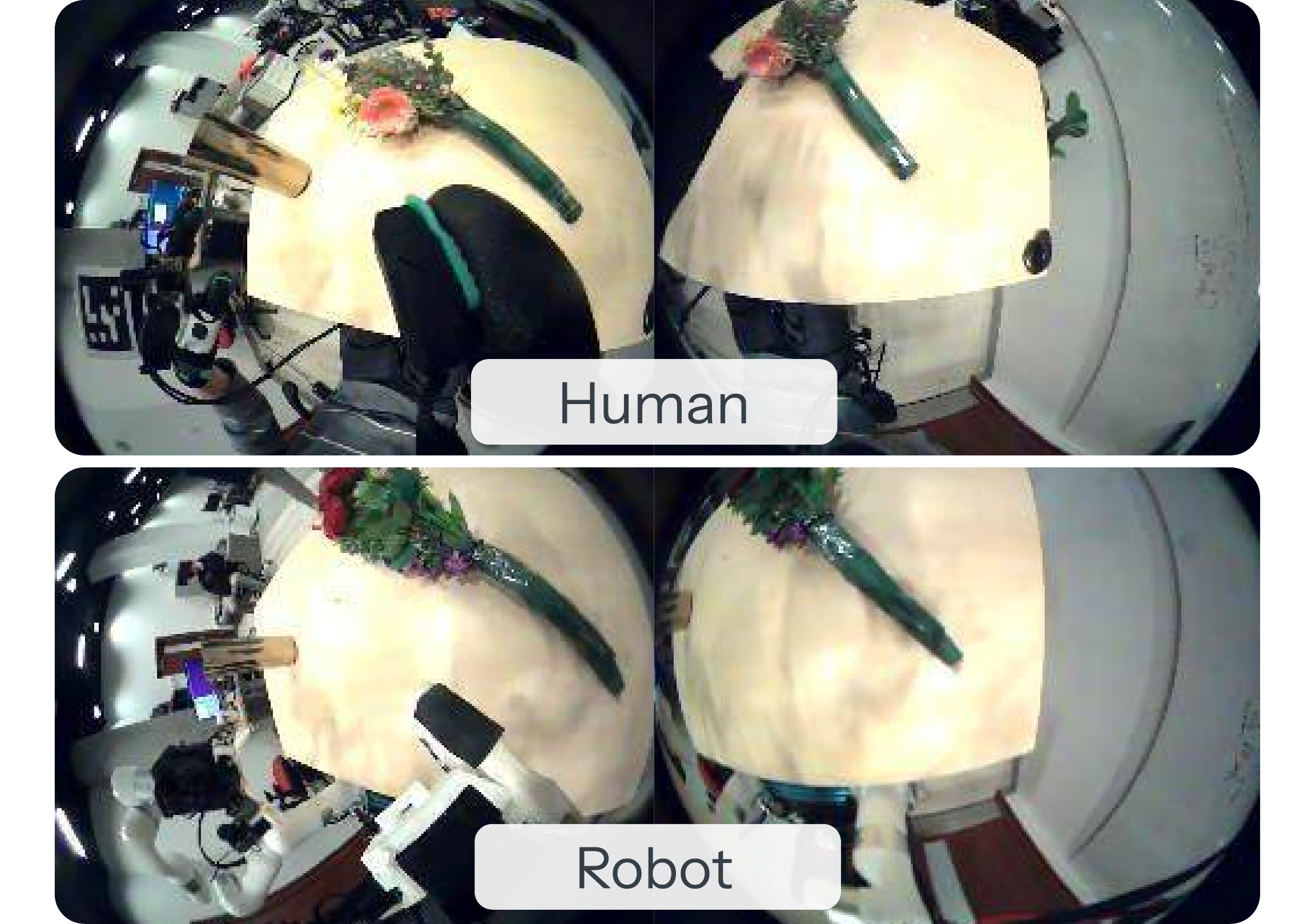}
\vspace{-0.1in}
  \caption{\small \our aligns the visual observations between humans and robots to bridge the embodiment gap. This incentivizes the model to learn a task-centric rather than embodiment-centric representation.}
 \label{fig:visualalignment}
\end{minipage}
\vspace{-0.2in}
\end{figure}

We begin by standardizing the observation space. Although our palm-mounted cameras have a wide field of view, we intentionally position them to focus primarily on the environment, minimizing the visibility of the hand itself. Importantly, the camera placement is mirrored between the human and robot hands. As shown in Figure~\ref{fig:visualalignment}, this design yields visually consistent observations across embodiments, allowing the policy to learn a shared visual representation that generalizes across both human and robot domains.

For action space alignment, we build on insights from prior work~\cite{handa2020dexpilot, robo-telekinesis}, optimizing robot hand kinematics to match the fingertip positions observed in human demonstrations. We note that this method is general and can work for any robot hand embodiment. It operates with fixed hyperparameters across users and is robust to variations in hand size—eliminating the need for user-specific tuning.

Collecting data using natural human hands offers benefits beyond ease of use. The diversity in hand morphology across human demonstrators introduces useful variation, which we hypothesize helps policies learn more generalizable grasping strategies—particularly important given the inherent mismatch between human and robot hand kinematics.

In summary, \our is a portable, high quality, human-centric system that can be worn by any operator to collect human data in real-world environments. Next, we explain how we use the data collected by \our to enable dexterous policies to generalize to in-the-wild scenarios.

\begin{figure*}
\centering
\includegraphics[width=\linewidth]{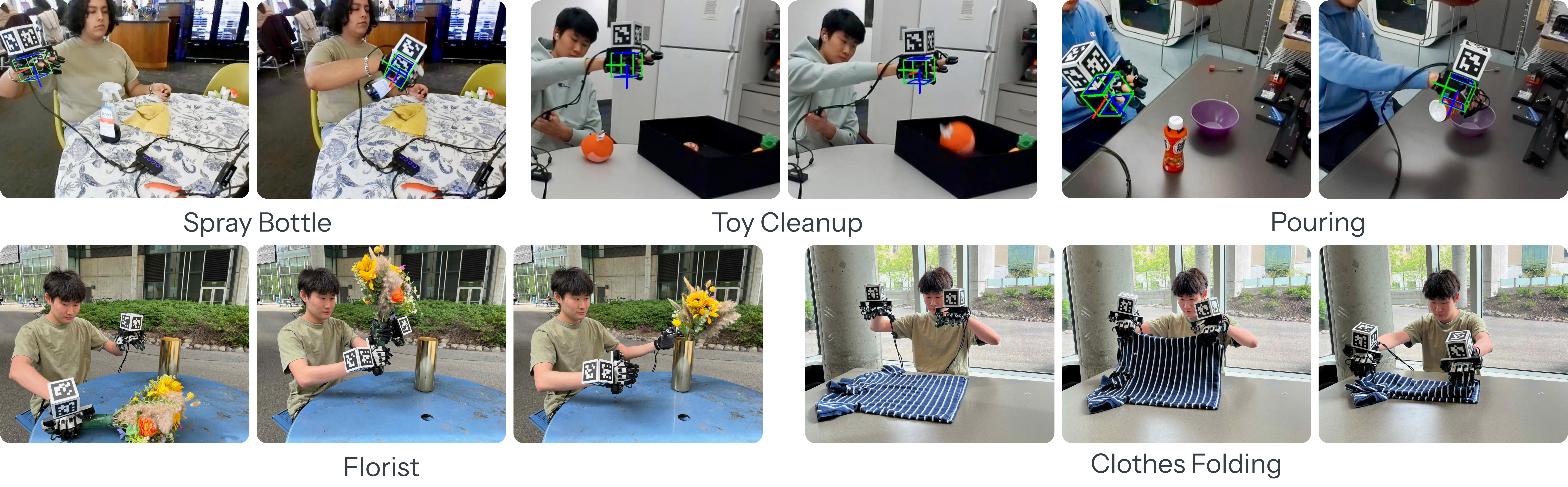}
\vspace{-0.2in}
  \caption{\small Using \oursystem, humans can effortlessly collect accurate data with their own hands across a wide range of environments. This data is directly used to train any robot hand to perform dexterous manipulation in a human-like way in any environment. We validate this approach on five representative tasks.  Please see videos of these tasks on our website at \url{https://dexwild.github.io}}
 \label{fig:tasks}
 \vspace{-0.1in}
\end{figure*}

\subsection{Training Data Modalities and Preprocessing}

Generalization in dexterous manipulation demands both scale and embodiment grounding. With this goal, \our collects two complementary datasets: a large-scale human demonstration dataset $D_H$ using \oursystem, and a smaller teleoperated robot dataset $D_R$. Human data offers broad task diversity and ease of collection in real-world settings, but lacks embodiment alignment. Robot data, while limited in scale, provides crucial grounding in the robot’s action and observation spaces. To harness the strengths of both, we co-train policies using a fixed ratio of human and robot data within a batch, $(w_h, w_r)$—balancing diversity with embodiment grounding to enable robust generalization during deployment.

At each training iteration, we sample a batch consisting of transitions ${x_h}$ and ${x_r}$ from $D_H$ and $D_R$, respectively, according to the co-training weights. Each transition $x_i$ at timestep $i$ contains:
\begin{itemize}
    \item \textbf{Observation} $o_i$: An observation at a given timestep consists of two synchronized palm camera images $I_{pinky}$ and $I_{thumb}$ captured at the current timestep, as well as a sequence of historical states, sampled at a step size up a given horizon $H$, comprising of $\{\Delta p_i, \Delta p_{i-\text{step}}, ..., \Delta p_{i-H}\}$. Each $\Delta p$ consists of relative historical end-effector positions. 
    \item \textbf{Action} $a_{i:i+n-1}$: An action chunk of size $n$ that includes actions $\{a_i, a_{i+1}, \dots, a_{i+n-1}\}$, where $a_i$ is the action at the current timestep. Specifically,  $a_i$ is a 26-dimensional vector consisting of:
    \begin{itemize}
        \item $a_{arm}$: A 9-dimensional vector describing relative end-effector position (3D) and orientation (6D).
        \item $a_{hand}$: A 17-dimensional vector describing the finger joint position targets of the robot hand.
    \end{itemize}
    For bimanual tasks, the observation and action spaces are duplicated, and the inter-hand pose is appended to the observation to facilitate coordination.
\end{itemize}

While our retargeting procedure brings human and robot trajectories into a shared action space, a few additional steps are necessary to make the human and robot datasets compatible for joint training:

\begin{itemize}
    \item \textbf{Action Normalization}: The actions of human and robot data are normalized separately to account for inherent distribution mismatches.
    \item \textbf{Demo Filtering}: Since human demonstrations are collected by untrained operators in uncontrolled environments, we apply a heuristic-based filtering pipeline to automatically detect and remove low-quality or invalid trajectories. This filtering step significantly improves dataset quality without manual labeling.
\end{itemize}

\subsection{Policy Training}

Through the careful design of our hardware, observation, and action interfaces, we are able to train dexterous robot policies using a simple behavior cloning (BC) objective~\cite{pomerleau1988alvinn,schaal1999imitation,ross2011reduction}. To effectively learn from our multimodal, diverse data, our training pipeline leverages large-scale pre-trained visual encoders and shows strong performance across different policy architectures. 

\textbf{Visual Encoder}: Training on DexWild data exposes our policy to significant visual diversity—across scenes, objects, and lighting—requiring an encoder that generalizes well to such variability. To address this, we adopt a pre-trained Vision Transformer (ViT) backbone, which has shown superior performance over ResNet-based encoders on in-the-wild manipulation tasks~\cite{ha2024umilegs, lin2024datascaling}. Pre-trained ViTs, especially those trained on large internet-scale datasets, are particularly effective at extracting rich, transferable features~\cite{nair2022r3m, Radosavovic2022, srirama2024hrp, dasari2023datasets}, making them well-suited for our setting.

\textbf{Policy Class}: While several imitation learning architectures have been proposed recently~\cite{zhao2023learning, chi2023diffusionpolicy}, we adopt a diffusion-based policy. Diffusion models are particularly well-suited for dexterous manipulation, as they can capture multi-modal action distributions more effectively than alternatives such as Gaussian Mixture Models (GMMs) or transformers. This capability becomes increasingly important in DexWild, where demonstrations are collected from multiple humans with diverse strategies, resulting in inherently multi-modal behaviors. As the dataset scales, modeling this variability becomes critical for robust policy learning. Specifically, \our uses a diffusion U-Net model~\cite{chi2023diffusionpolicy} to generate action chunks.

Concretely, the training procedure is outlined in Algorithm~\ref{alg:policy_training}.
\begin{algorithm}[H]
\caption{\small \our Imitation Learning Procedure}\label{alg:policy_training}
\begin{algorithmic}[1]
\small
\REQUIRE Human dataset $\mathcal{D}_H$, Robot dataset $\mathcal{D}_R$, Co-training weights $\{\omega_h, \omega_r\}$
\STATE Initialize policy $\pi_\theta$ with ViT encoder $\phi_\text{vit}$
\WHILE{not converged}
    \STATE Sample a batch of transitions $\{x_h\}, \{x_r\}$ from $\mathcal{D}_H, \mathcal{D}_R$ using weights $\{\omega_h, \omega_r\}$
    \FOR{each transition $x_i$ in the batch}
        \STATE Extract observation $o_i$
        \STATE Encode images: $Z_i = \phi_\text{vit}(o_i)$
        \STATE Extract ground truth action chunk $a_{i:i+n-1} = \{a_i, \dots, a_{i+n-1}\}$
        \STATE Sample noise scale $t \sim \mathcal{U}(1, T)$
        \STATE Add noise $\epsilon_t \sim \mathcal{N}(0, \sigma_t)$ to $a_{i:i+n-1}$
        \STATE Predict noise $\hat{\epsilon}_\theta = \pi_\theta(Z_i,  a_{i:i+n-1} + \epsilon_t, t)$
        \STATE Compute diffusion loss $\mathcal{L}_\theta = \|\epsilon_t - \hat{\epsilon}_\theta\|_2^2$
    \ENDFOR
    \STATE Update policy parameters $\theta$
\ENDWHILE
\end{algorithmic}
\end{algorithm}

\begin{figure*}
\centering
\includegraphics[width=\linewidth]{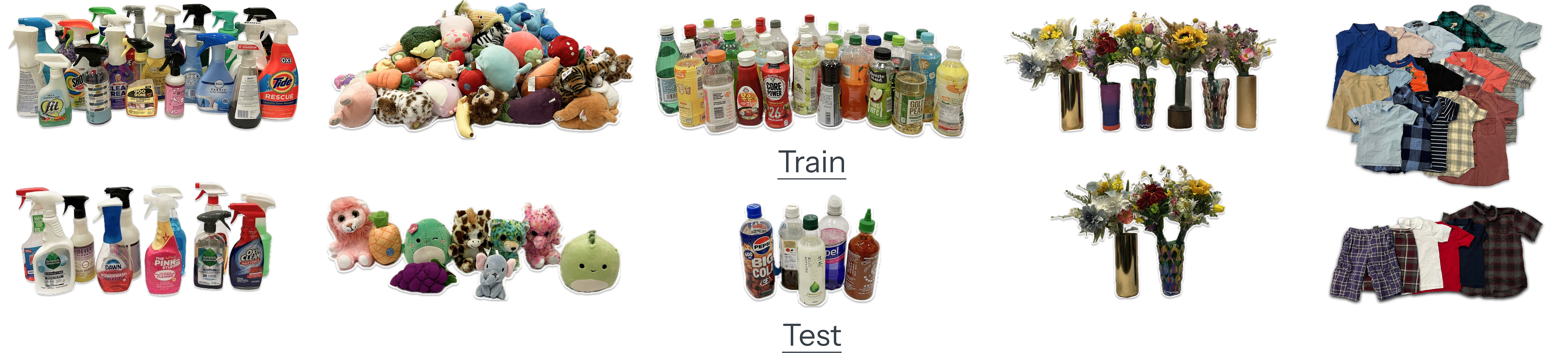}
\vspace{-0.1in}
  \caption{\small We collect data using a diverse set of objects across categories. \textit{Spray Bottle Task} – 25 Train, 11 Test; \textit{Toy Cleanup Task} – 64 Train, 9 Test; \textit{Pour Task} – 35 Train, 5 Test; \textit{Florist Task} - 6 Train, 2 Test; \textit{Clothes Folding Task} - 17 Train, 6 Test.}
 \label{fig:traintestobjects}
\vspace{-0.2in}
\end{figure*}

An important finding in our training framework is that tuning the human-to-robot data weighting significantly affects real-world performance. We discuss these effects in Section~\ref{sec:results-main}.

\section{Experiments}
Our experimental evaluation encompasses extensive real-world deployment across diverse environments and robots, utilizing both human demonstrations and robot teleoperation data. Below, we outline our data collection process, experimental setup, and evaluation tasks.

\subsection{Scaling up Data Collection}
Our hardware system was deployed to 10 untrained users to collect data across a wide range of real-world environments. These settings included indoor and outdoor locations, day and night conditions, crowded cafeterias and quiet study areas, with varied tables, objects, and lighting setups. The collectors themselves varied in hand sizes and demonstration styles, enabling us to learn from a wide distribution of environments and interactions.

We constructed two datasets through our collection efforts: $D_H$ (human-collected data) and $D_R$ (robot-collected data). The human dataset $D_H$ comprises 9,290 demonstrations across five tasks: 3,000 demonstrations from 30 different environments for each of the \textit{Spray Bottle} and \textit{Toy Cleanup} tasks, 621 trajectories from 6 environments for the \textit{Pour} task, 1,545 demonstrations from 15 environments for the \textit{Florist} task, and 1,124 demonstrations from 12 environments for the \textit{Clothes Folding} task.

The robot dataset $D_R$ includes 1,395 demonstrations: 388 for \textit{Spray Bottle}, 370 for \textit{Toy Cleanup}, 111 for \textit{Pour}, 236 for \textit{Florist}, and 290 for \textit{Clothes Folding} tasks. Robot data was collected using an xArm and LEAP hand V2 Advanced. Our training and test objects are detailed in Figure \ref{fig:traintestobjects}. 

\subsection{Evaluation Tasks}

We evaluate our approach on five diverse manipulation tasks, each designed to assess specific aspects of dexterous manipulation: functional grasping, long-horizon planning, cross-task transfer, bimanual coordination, and deformable object manipulation. A task visualization is provided in Figure \ref{fig:tasks}. 

In the Spray Bottle task, the robot grasps a spray bottle by the handle and sprays a target cloth, testing functional grasping and affordance understanding. In Toy Cleanup, the robot picks up scattered toys and places them in a bin, evaluating generalization and long-horizon planning. The Pouring task involves tilting a bottle to pour into a container, demonstrating skill transfer from the spray bottle task. In Bimanual Florist, the robot hands over a flower between its arms and inserts it into a vase, testing precise bimanual coordination. Finally, in Bimanual Clothes Folding, the robot uses both hands to fold a clothing item, assessing manipulation of deformable objects. Full task specifications and scoring criteria for all tasks are provided in Appendix~\ref{app:taskspec}.

These tasks systematically evaluate \ours \textit{functional grasping} capabilities, \textit{generalization} across object types, \textit{transferal} of skills across tasks, \textit{coordination} between arms, and \textit{adaptability} to deformable objects. Success requires the policy to adapt to varying object properties, environmental conditions, and task constraints. 
\begin{figure*}
\centering
\includegraphics[width=\linewidth]{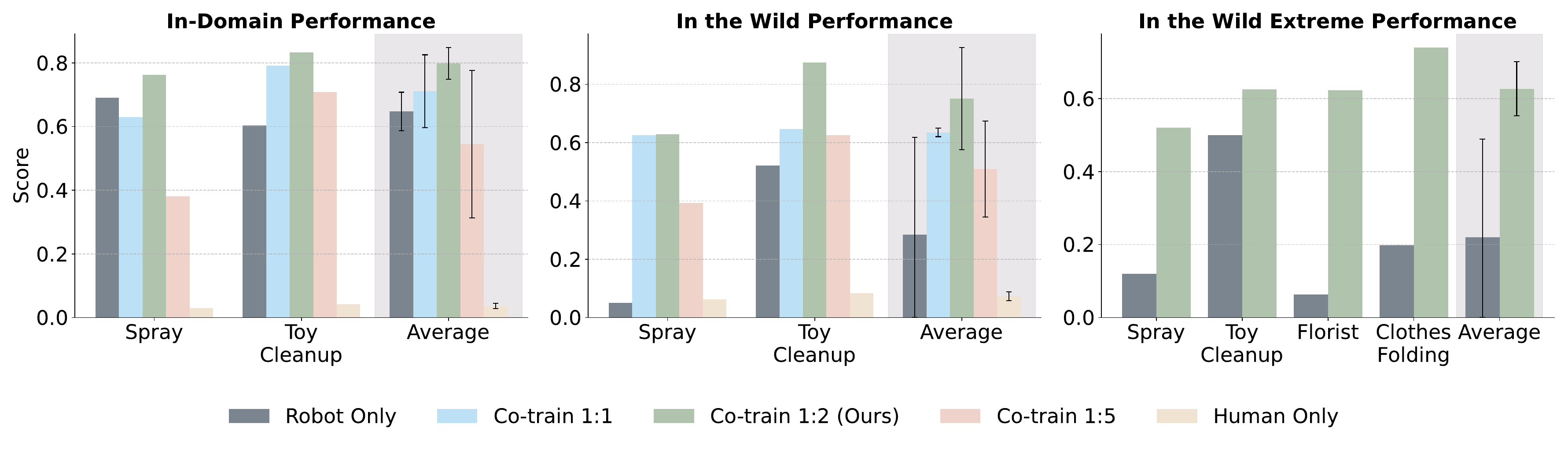}
\vspace{-0.2in}
  \caption{\small \textbf{How does co-training help with scaling up in the wild performance?}
  We evaluate our policy across three scenarios: (a) In-Domain scenes where robot training data was collected but with novel objects, (b) In-the-Wild scenes present in DexWild but not in robot data, and (c) In-the-Wild Extreme scenes absent from both datasets. Displayed ratio is Robot:Human.}
 \label{fig:main-cotrainperf}
 \vspace{-0.2in}
\end{figure*}

\subsection{Evaluation Environments}
For robot experiments, we employed an xArm robot and Franka system, both equipped with either LEAP hand or LEAP hand V2 Advanced \cite{shaw2023leaphand, shaw2024bimanual}. Unless explicitly mentioned, xArm and LEAP hand V2 Advanced was used. We evaluate our approach across three scenarios:
\begin{enumerate}
    \item In-Domain: Environments where robot training data was collected, testing with novel objects
    \item In-the-Wild: Environments present in DexWild but absent from robot training data
    \item In-the-Wild Extreme: Unseen environments absent from both datasets. 
\end{enumerate}

\vspace{-0.1in}
\section{Analysis and Results}

In our evaluations, we seek to investigate the following key questions:

\begin{enumerate}
\item How effectively does \our leverage human data to achieve strong in-the-wild performance?

\item Does \our enable policy transfer across tasks and robot embodiments?

\item Does policy performance scale effectively with increasing amounts of \oursystem data?

\end{enumerate}
Please see videos of our results at \url{https://dexwild.github.io}.
\subsection{Zero Shot In the Wild Policies w/ \our}
\label{sec:results-main}

\textbf{\our enables strong policy generalization in novel scenes.} We evaluate policies in environments with increasing novelty to assess their generalization. As shown in Figure \ref{fig:main-cotrainperf}, policies trained exclusively on robot data perform well in in-domain settings (64.7\% success rate) but degrade significantly in more challenging scenarios—in-the-wild (28.5\%) and in-the-wild extreme (22.0\%). This 36-point performance drop suggests that robot-only policies overfit to environment-specific features and fail to develop robust, transferable representations. In contrast, policies trained only on human data learn high-level object affordances and approach objects reliably, even in complex scenes. However, without robot-specific action grounding, they struggle to execute precise manipulation, resulting in poor performance across all scenarios (3.6\% in-domain, 7.3\% in-the-wild). 

To combine the strengths of both modalities, we adopt a co-training strategy—jointly training on both robot and human data—a method validated in prior works~\cite{open_x_embodiment_rt_x_2023, om2023octo, khazatsky2024droid, kareer2024egomimic, qiu2025-humanpolicy}. This encourages the policy to learn task-relevant features rather than overfitting to specific embodiments or environments. We experiment with different \textbf{robot-to-human} data ratios (1:1 to 1:5) per training batch. Our empirical analysis reveals that a 1:2 ratio yields optimal performance across all scenarios:

\begin{enumerate} 
\item In Domain: 79.8\% vs. 64.7\% (robot-only)
\item In-the-wild: 75.1\% vs. 28.5\% (robot-only)
\item In-the-wild Extreme: 62.7\% vs. 22.0\% (robot-only)
\end{enumerate}

Interestingly, increasing the human data ratio further (e.g., 1:5) degrades performance (54.5\% in-domain, 50.9\% in-the-wild), indicating that robot data remains essential for grounding fine-grained control.

\textbf{\our extends to complex bimanual coordination tasks.} To evaluate whether \our generalizes beyond single-arm tasks, we test it on bimanual tasks that demand precise coordination between two hands. We compare co-trained policies (1:2 ratio) against robot-only policies in in-the-wild extreme settings. \our policies achieve a strong 68.1\% average success rate, compared to just 13\% for the robot-only baseline. Even when failures occur, \our policies exhibit meaningful attempts at task execution—while robot-only policies often produce erratic or unstructured behavior. 

These results demonstrate that \our not only enables robust generalization across environments but also scales to more complex manipulation behaviors.

\subsection{Robust Cross-Task and Cross-Embodiment Generalization}

\textbf{\our enables transfer of low-level skills across tasks.} 
Many manipulation tasks share foundational motor skills—such as lifting, orienting, and rotating objects—which opens the door to skill reuse across related tasks. For example, opening a microwave and opening a cupboard both involve similar coordination and control. We evaluate this form of cross-task transfer using the \textit{pouring} task, which shares many motion primitives with the \textit{spray} task. Crucially, we use no robot data for pouring and instead combine human (\oursystem) demonstrations of pouring with robot demonstrations from spraying. This setup enables \textbf{zero-shot generalization} to pouring in in-the-wild extreme environments. Using a 1:2 robot-to-human co-training ratio, our policy achieves a \textbf{94\% success rate}, far exceeding policies trained with only robot (0\%) or only human data (11\%).

\textbf{\our enables transfer across robot embodiments.}  
Since \our data is not tied to any specific embodiment, it naturally supports cross-platform transfer. This prolongs the value of our data, as collecting platform-specific data for every new robot is resource-intensive and impractical. We test two transfer scenarios in in-the-wild extreme scenes:
\begin{itemize}
    \item \textbf{Cross-arm}: Transferring from an xArm to a Franka Panda arm. We achieve a 37.5\% success rate, compared to 4.5\% for the robot-only baseline—an \textbf{8.3$\times$ improvement}.
    \item \textbf{Cross-hand}: Transferring from the LEAP Hand V2 Advanced to the original LEAP Hand. We achieve 65.3\% success versus 13.3\% for the baseline, showing that \our generalizes not only across arms, but across dexterous hands as well.
\end{itemize}

\begin{figure*}
\begin{minipage}[t]{1.0\textwidth} 
\centering
\includegraphics[width=\linewidth]{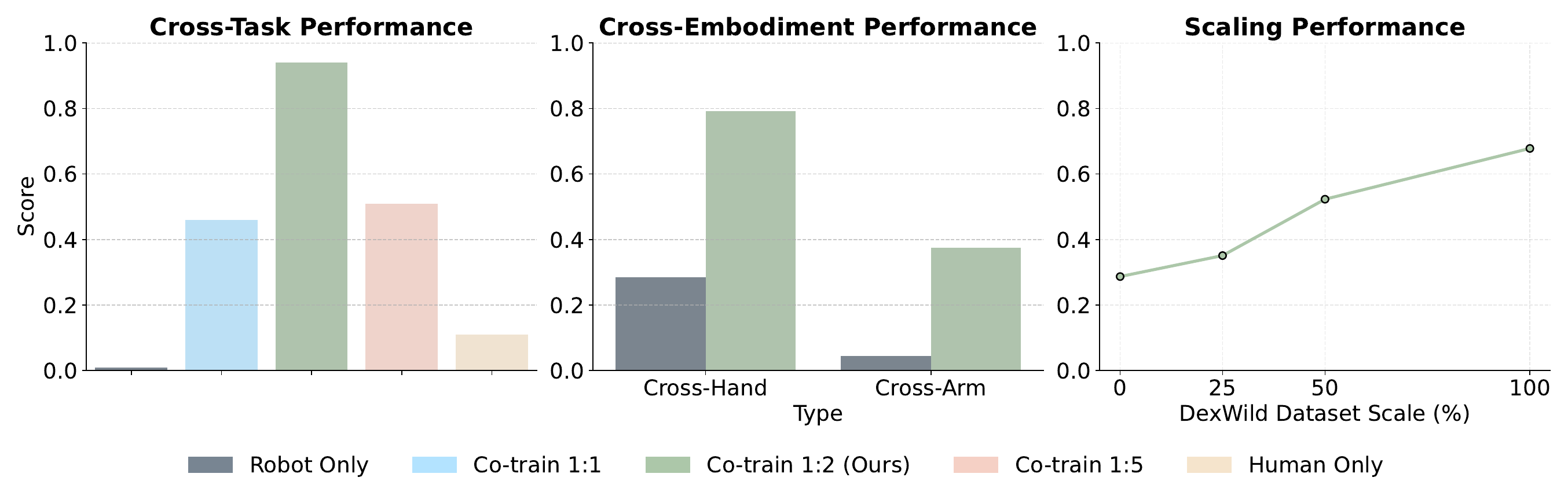}
\vspace{-0.2in}
\caption{\small Left: \textbf{Cross-Task Performance} – Evaluating \our on the pour task using robot data exclusively from the spray task. Middle: \textbf{Cross-Embodiment Performance} – Testing \our policy on the Original LEAP hand and a Franka robot arm. Right: \textbf{Scaling Performance} – Demonstrating improved \our performance as dataset size increases. Displayed ratio is Robot:Human.}
 \label{fig:main-crossperf}
\end{minipage}
\vspace{-0.2in}
\end{figure*}

These results, shown in Figure~\ref{fig:main-crossperf}, demonstrate that \our enables zero-shot generalization to new tasks and hardware embodiments \textbf{without any additional robot-specific data}, making it an efficient and general framework for dexterous policy learning on many robots.

\begin{figure}[b]
\begin{minipage}[t]{0.49\textwidth} 
\centering
\includegraphics[width=\linewidth]{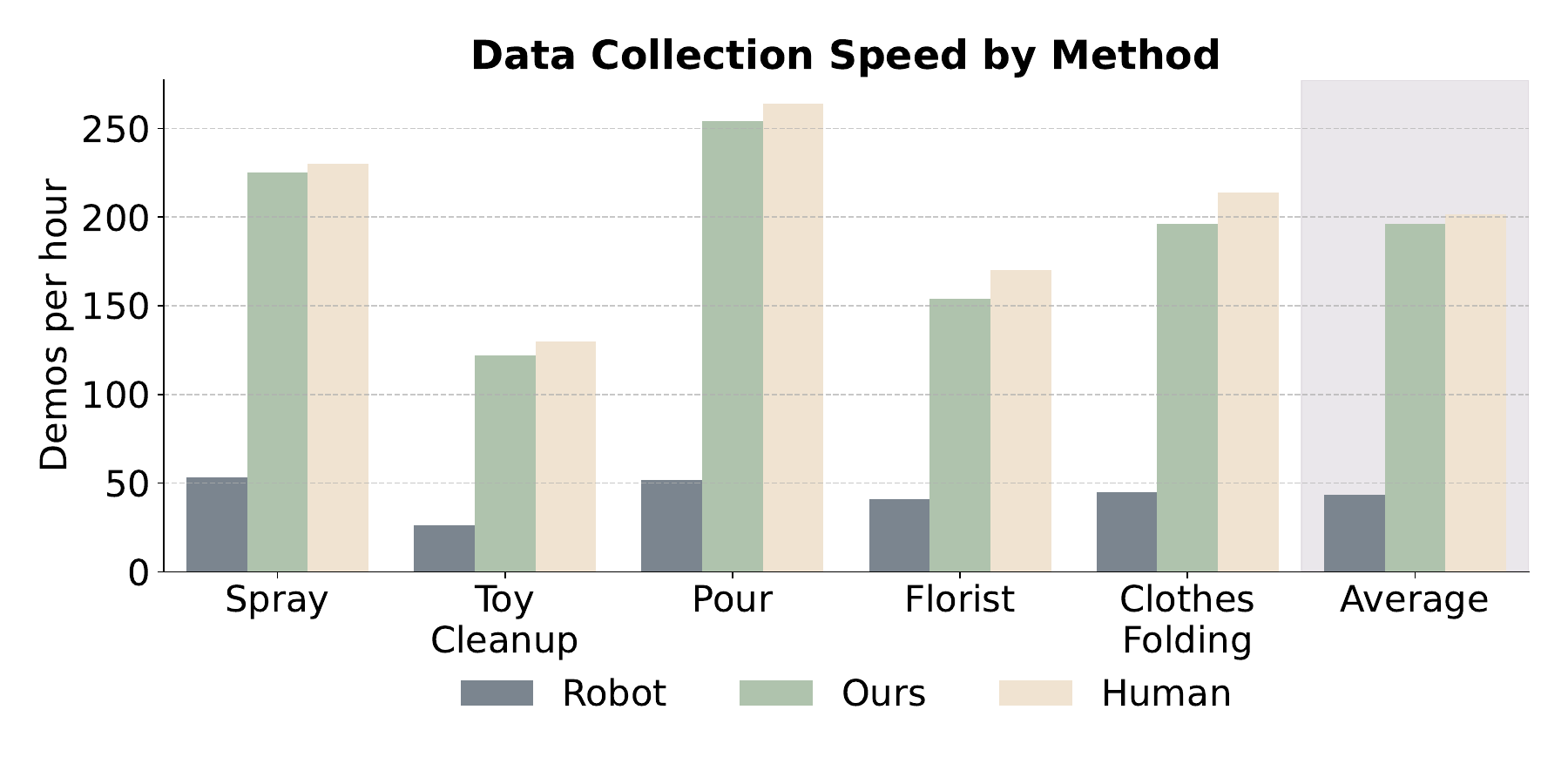}
\vspace{-0.3in}
  \caption{\small \oursystem offers \textbf{4.6$\times$} improvement over robot data collection speed and nearly matches the human bare hands data collection speed.}
 \label{fig:datacollectionperf}
\end{minipage}
\vspace{-0.2in}
\end{figure}

\subsection{Scalability of \our}

\textbf{Policy performance scales with dataset size.}
To understand how data scale impacts policy performance in the wild, we randomly sample subsets of the full human dataset at varying sizes and evaluate the resulting policies. We fix the size of the robot dataset. As shown in Figure~\ref{fig:main-crossperf}, there is a clear positive correlation between dataset size and average task performance—rising from 28.7\% at 20\% dataset size to 67.8\% with the full dataset, marking a 2.36$\times$ improvement. Interestingly, the learning curve is nonlinear, with especially steep gains in the 25–50\% range, suggesting a critical threshold where the policy begins to reliably learn generalizable behaviors.

Importantly, performance continues to improve all the way to 100\% data usage, indicating that the system has not yet plateaued. This suggests that even more capable policies could be learned with continued data collection.

\textbf{\oursystem enables fast and scalable data collection.}  
Given the observed benefits of scaling, we evaluate the data collection efficiency of \oursystem via a comparative user study measuring demonstrations per hour. As shown in Figure~\ref{fig:datacollectionperf}, \oursystem achieves an average collection rate of \textbf{201 demos/hour} across five representative tasks—nearly matching the rate of demonstrations collected using bare hands and \textbf{4.6$\times$ faster} than a traditional robot teleoperation system based on Gello~\cite{shaw2024bimanual, wu2023gello}, which achieves just 43 demos/hour.

We identify three key limitations of Gello-based collection that our system overcomes:
\begin{enumerate}
    \item \textbf{Lack of haptic feedback:} Operators cannot feel objects, making fine manipulation difficult for certain tasks.
    \item \textbf{Scene reset:} Resetting the environment is cumbersome and often requires a second operator or pauses in data collection.
    \item \textbf{Hardware setup overhead:} Robots are heavy and require time-consuming setup at each new location, whereas \oursystem is portable and can be set up in minutes.
\end{enumerate}

\our not only demonstrates strong scaling trends with increasing data volume, but also offers a practical and efficient path to collecting diverse, high-quality data at scale—crucial for real-world generalization.

\label{sec:conclusion}
\section{Conclusion and Limitations}
We introduce \our, a scalable framework for learning dexterous manipulation policies that effectively generalize to new tasks, environments, and robot embodiments. We introduce \oursystem, a portable, human-centric data collection device that significantly accelerates dataset creation (4.6$\times$ faster than conventional robot teleoperation). We propose \our cotraining method, which leverages large scale human demonstrations alongside minimal robot data to achieve robust generalization-reaching a success rate of 68.5\% in completely unseen environments, nearly four times higher than methods using robot data only. Furthermore, \our's embodiment-agnostic design enables strong cross-embodiment and cross-task transfer capabilities, reducing the need for robot-specific data.

Despite these strengths, several limitations remain that motivate future research:
First, our approach still depends on a limited number of teleoperated robot data to bridge the gap between human and robot actions. Future work could explore improved retargeting or online policy adaptation to remove the need for teleoperated data. Next, because humans typically perform these tasks successfully, their demonstrations seldom include error recovery—causing trained policies to struggle to recover from unexpected failures. Adding recovery examples or adaptive strategies could boost real-world robustness.  Finally, our method uses only visual and kinematic data, which limits its performance in contact-rich tasks. Incorporating tactile or haptic sensing could improve the handling of delicate interactions.

In summary, \our represents a significant step toward scalable, generalizable robot manipulation policies. Our results highlight the promise of leveraging human interaction data at scale, offering an exciting avenue toward truly dexterous and versatile robots operating in diverse, real-world environments.

Videos, code, and hardware instructions are available on our website at \url{https://dexwild.github.io}

\clearpage

\section*{Acknowledgments}
We would like to thank Yulong Li, Hengkai Pan, and Sandeep Routray for thoughtful discussions.  We'd also like to thank Andrew Wang for setting up compute and Yulong Li for helping with robot system setup. Lastly, we'd like to express thanks to Hengkai Pan, Andrew Wang, Adam Kan, Ray Liu, Mingxuan Li, Lukas Vargas, Jose German, Laya Satish,
Sri Shasanka Madduri for helping collect data.  This work was supported in part by AFOSR FA9550-23-1-0747 and Apple Research Award.



\clearpage
\section{Appendix}

Videos of our results, code to recreate our system, and hardware instructions are available on our website at \url{https://dexwild.github.io}

\subsection{Detailed Task Description and Scoring Criteria:}
\label{app:taskspec}

We evaluate five dexterous manipulation tasks, each designed to assess different capabilities such as functional grasping, long-horizon planning, precision, bimanual coordination, and deformable object manipulation. Each task is scored according to a structured rubric based on discrete completion milestones.

The task scoring criteria are designed to quantify the performance of different robot tasks based on specific completion milestones. Each task has a set of defined actions with corresponding point values. Higher scores are assigned to more complex or functionally successful actions, while partial completions and failed attempts receive lower scores. This structured scoring system allows for consistent evaluation and comparison of task performance.

\vspace{1mm}
\noindent \textbf{Spray Bottle} \\
This task evaluates functional grasping and affordance understanding. The robot must grasp a spray bottle and orient it to spray over a target cloth.
\begin{itemize}
    \item[--] 0.00: Nothing 
    \item[--] 0.15: Tries functional grasp but fails
    \item[--] 0.25: Grasp bottle 
    \item[--] 0.75: Grasp bottle, orient over cloth 
    \item[--] 0.75: Grasp bottle, use functional grasp 
    \item[--] 1.00: Grasp bottle, use functional grasp, orient over cloth
\end{itemize}

\vspace{1mm}
\noindent \textbf{Toy Cleanup} \\
This task tests long-horizon planning and generalization. The robot must collect scattered toys and deposit them in a designated bin.
\begin{itemize}
    \item[--] 0.00: Nothing 
    \item[--] 0.25: Tries for grasp but fails
    \item[--] 0.50: Grasp object 
    \item[--] 1.00: Grasp object, drop into bin
\end{itemize}

\vspace{1mm}
\noindent \textbf{Pouring} \\
This task assesses precise motion control and transfer learning from the spray bottle task. The robot must pour liquid from a bottle into a container.
\begin{itemize}
    \item[--] 0.00: Nothing 
    \item[--] 0.15: Tries functional grasp but fails 
    \item[--] 0.25: Grasp bottle 
    \item[--] 0.75: Grasp bottle, pour into container 
    \item[--] 0.75: Grasp bottle, use functional grasp 
    \item[--] 1.00: Grasp bottle, use functional grasp, pour into container
\end{itemize}

\vspace{1mm}
\noindent \textbf{Bimanual Florist} \\
This task evaluates coordinated control of both hands. The robot must pick up a flower, hand it to the other arm, and insert it into a vase.
\begin{itemize}
    \item[--] 0.00: Nothing 
    \item[--] 0.15: Tries grasp but fails 
    \item[--] 0.25: Grasp the bouquet 
    \item[--] 0.75: Grasp the bouquet, handover 
    \item[--] 1.00: Grasp the bouquet, handover, insert into vase
\end{itemize}

\begin{figure}[t]
\centering
\includegraphics[width=0.95\linewidth]{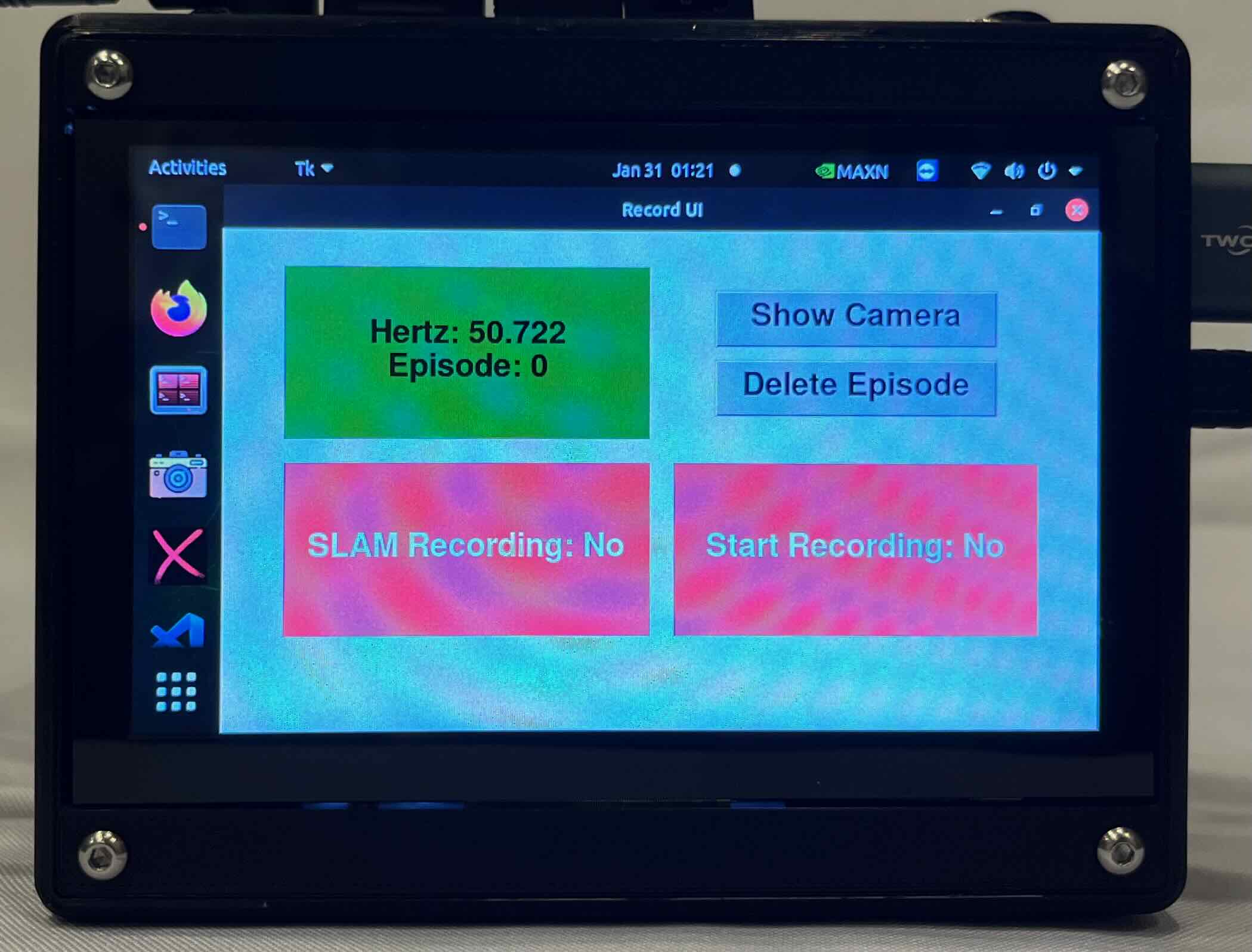}
\caption{\small \oursystem features a simple and easy-to-use interface for deployment by untrained data collectors.}
\label{fig:datacollection_interface}
\vspace{-0.1in}
\end{figure}

\vspace{1mm}
\noindent \textbf{Clothes Folding} \\
This task tests manipulation of deformable objects using both hands. The robot must fold a clothing item placed on a surface.
\begin{itemize}
    \item[--] 0.00: Nothing 
    \item[--] 0.25: Tries grasp but fails 
    \item[--] 0.50: Grasp with one hand 
    \item[--] 0.75: Grasp with both hands 
    \item[--] 1.00: Grasp and fold
\end{itemize}

\subsection{Data Collection Procedure}
To deploy \oursystem with untrained data collectors, we provide a one-page instruction sheet outlining the task, object setup, and system startup/shutdown. \oursystem includes three core components: a wrist-tracking camera, a battery-powered mini-PC for onboard data capture, and a custom sensor pod with a motion-capture glove and palm-mounted cameras. At a new site, users simply wear the mocap glove and power on the mini-PC with a provided power bank. For egocentric tracking, a headstrap holds the tracking camera; for exocentric tracking, we provide a collapsible tripod. Once booted, users launch our custom desktop app and control recording via a Bluetooth clicker or foot pedal. The UI (Fig.~\ref{fig:datacollection_interface}) shows sensor status, SLAM recording, and data capture indicators, along with buttons to view the tracking camera feed and delete the last episode. Collectors gather 100 episodes per location. After the day is finished, we upload the data to our remote machine for processing.

\subsection{Downstream Data Processing}
Each episode is stored in its own folder, with subfolders organizing individual actions and observations. SVO recordings from the Zed Mini camera—used for SLAM and wrist pose tracking—are saved separately, with each file covering five episodes. To begin data processing, we use the Zed SDK to decode these SVO files, reconstruct the camera’s motion, and perform ArUco cube tracking and wrist pose estimation using both the left image and stereo depth data. We then apply a filtering pipeline to assess tracking quality; episodes are discarded if the wrist pose cannot be reliably tracked for more than 75\% of the duration. Next, we compute the action distribution and clip outliers outside the 2nd and 97th percentiles. We smooth the trajectories using interpolation and Gaussian filtering to ensure fluid motion. Hand motions are then retargeted using inverse kinematics in PyBullet, following the method in \cite{shaw2024bimanual}. The entire pipeline is parallelized using Ray for efficiency.

\subsection{Behavior Cloning Policy Architecture and Training Hyper-Parameters}
\label{app:bc_params}

Our behavior cloning policy takes as input RGB images and relative state history. We obtain tokens for the image observation via a ViT and tokens for relative states via linear layers. The weights of ViT is initialized from the Soup 1M model from~\cite{dasari2023datasets}. We decide to include relative states as we found it greatly increases the robustness of the policy, and enables smoother motions. In particular, for bimanual tasks, we find that including the interhand pose (pose of left hand relative to right hand) greatly increases success rate in tasks like Florist We implement both Action Chunking Transformer \cite{zhao2023learning} and Diffusion U-Net \cite{chi2023diffusionpolicy} as policy classes, which output a sequence of actions. The network outputs actions which consists of relative end effector actions and absolute hand joint angles.

We list the hyper-paramaters that we used for policy training using behavior-cloning in this Table \ref{tab:full-hyperparams}

\subsection{Low Level Motion Control}
 For optimal smoothness of our policies and safety, we employ a Riemannian Motion Policy (RMP) \cite{ratliff2018riemannianmotionpolicies} implemented in Isaac Lab \cite{Mittal_2023}, where the RMP dynamically generates joint-space targets given end effector targets. RMP also has the added benefit of incorporating real-time collision avoidance, preventing self-collision between the arms and a set table height. Although our policies does not rely on RMP to prevent collisions, the peace of mind is appreciated.

\subsection{Comparing Policy Classes}
\textbf{Does \our work with different behavior cloning policy classes?} Table~\ref{tab:act_diffusion} compares the performance of ACT and Diffusion—across both the In-the-Wild and In-the-Wild Extreme settings. Each policy is evaluated in a robot-only setting and a co-trained (1:2) setting using the DexWild dataset. Notably, Diffusion policies benefit more from DexWild co-training, achieving the highest scores in all tasks, including substantial improvements on the Pour task where the policy must generalize across tasks. These results suggest that DexWild co-training enables stronger generalization, especially when paired with expressive policy architectures like Diffusion.

\subsection{Cross Hand Extended Results}
\label{app:crosshand}
\textbf{Does \our generalize across different robot hands?}  
Table \ref{tab:leap_v1_results} reports LEAP Hand performance under both \emph{In the Wild} and \emph{In the Wild Extreme} conditions.  In every case, DexWild co-training substantially outperforms the robot-only baseline. These results highlight the effectiveness of DexWild in cross embodiment generalization even when using a completely different robot hand.

\subsection{Scaling Extended Results}
\label{app:scaling}
\textbf{Does \our improve as more \our data is added?}  
Table~\ref{tab:scaling_results} shows steady gains as we scale from 0\% to 100\% of the DexWild dataset. Performance increases steadily with more human demonstrations, with a notable jump between 25\% and 50\% of the dataset. These results demonstrate that \our enables scalable learning, where even comparably smaller data scales yields substantial gains, and additional data continues to enhance generalization

\subsection{Cotraining Extended Results}
\label{app:cotraining}
\textbf{How does \our react to different cotraining ratios?}  
Table~\ref{tab:cotrain_grouped} groups all three raw metrics: (a) In-Domain, (b) In-the-Wild, and (c) In-the-Wild Extreme. All evaluations were run on xArm + LEAP Hand V2 Advanced.

\vspace{0.7in}
\begin{table}[!htb]
    \vspace{0em}
    \centering
    \resizebox{\columnwidth}{!}{%
    \begin{tabular}{l l c c c c}
    \toprule
    \textbf{Task} & \textbf{Policy Class} & \multicolumn{2}{c}{\textbf{In the Wild}} & \multicolumn{2}{c}{\textbf{In the Wild Extreme}} \\
    \cmidrule(lr){3-4} \cmidrule(lr){5-6}
    & & \textbf{Robot Only} & \textbf{1:2} & \textbf{Robot Only} & \textbf{1:2} \\
    \midrule

    \multirow{2}{*}{Spray}
    & ACT        & 0.000 & \textbf{0.680} & 0.115 & \textbf{0.395} \\
    & Diffusion  & 0.050 & \textbf{0.628} & 0.120 & \textbf{0.520} \\
    \midrule

    \multirow{2}{*}{Toy Cleanup}
    & ACT        & 0.458 & \textbf{0.583} & 0.125 & \textbf{0.458} \\
    & Diffusion  & 0.521 & \textbf{0.875} & 0.500 & \textbf{0.625}\\
    \midrule

    \multirow{2}{*}{Pour (Cross Task)}
    & ACT        & 0.025 & \textbf{0.508} & 0.000 &  \textbf{0.350}  \\
    & Diffusion  & 0.000 & \textbf{0.958} & 0.000 &  \textbf{0.917}  \\
    \bottomrule
    \end{tabular}
    }
    \caption{DexWild Performance on Different Policy Classes}
    \label{tab:act_diffusion}
    \vspace{-1em}  
\end{table}

\begin{table}[H]
  \centering
  \resizebox{\columnwidth}{!}{%
  \begin{tabular}{lcc|cc}
    \toprule
    & \multicolumn{2}{c}{\textbf{In the Wild}} 
    & \multicolumn{2}{c}{\textbf{In the Wild Extreme}} \\
    \cmidrule(lr){2-3}\cmidrule(lr){4-5}
    \textbf{Task}          & \textbf{Robot Only} & \textbf{1:2} 
                           & \textbf{Robot Only} & \textbf{1:2} \\
    \midrule
    Spray                  & 0.305 & \textbf{0.805} & 0.150 & \textbf{0.600} \\
    Toy Cleanup            & 0.500 & \textbf{0.656} & 0.250 & \textbf{0.542} \\
    Pour (Cross Task)      & 0.050 & \textbf{0.917} & 0.000 & \textbf{0.817} \\
    \bottomrule
  \end{tabular}
  }
  \caption{LEAP Hand Performance on In-the-Wild and In-the-Wild Extreme Tasks. Ratio is Robot:Human}
  \label{tab:leap_v1_results}
\end{table}

\begin{table}[H]
  \centering
  \begin{tabular}{lcccc}
    \toprule
    \textbf{Scale} & \textbf{0\%} & \textbf{25\%} & \textbf{50\%} & \textbf{100\%} \\
    \midrule
    Spray        & 0.060 & 0.260 & 0.605 & 0.565 \\
    Toy Cleanup  & 0.514 & 0.442 & 0.440 & 0.792 \\
    \midrule
    \textbf{Average} & \textbf{0.287} & \textbf{0.351} & \textbf{0.523} & \textbf{0.678} \\
    \textbf{Std}     & \textbf{0.321} & \textbf{0.129} & \textbf{0.116} & \textbf{0.160} \\
    \bottomrule
  \end{tabular}
  \caption{Performance Scaling with DexWild Dataset Size}
  \label{tab:scaling_results}
\end{table}

\begin{table}[H]
  \centering
  \begin{subtable}[t]{\linewidth}
    \centering
    \begin{tabular}{lccccc}
      \toprule
      \textbf{Task} & \textbf{Robot} & \textbf{1:1} & \textbf{1:2} & \textbf{1:5} & \textbf{Human} \\
      \midrule
      Spray       & 0.690 & 0.630 & 0.763 & 0.381 & 0.030 \\
      Toy Cleanup & 0.604 & 0.792 & 0.833 & 0.708 & 0.042 \\
      \midrule
      \textbf{Average} & \textbf{0.647} & \textbf{0.711} & \textbf{0.798} & \textbf{0.545} & \textbf{0.036} \\
      \textbf{Std}     & \textbf{0.061} & \textbf{0.114} & \textbf{0.050} & \textbf{0.232} & \textbf{0.008} \\
      \bottomrule
    \end{tabular}
   \caption{In Distribution Task Performance}
  \end{subtable}
 
  \begin{subtable}[t]{\linewidth}
    \centering
     \vspace{5mm}
    
    \begin{tabular}{lccccc}
      \toprule
      \textbf{Task} & \textbf{Robot} & \textbf{1:1} & \textbf{1:2} & \textbf{1:5} & \textbf{Human} \\
      \midrule
      Spray       & 0.050 & 0.625 & 0.628 & 0.393 & 0.063 \\
      Toy Cleanup & 0.521 & 0.646 & 0.875 & 0.625 & 0.083 \\
      \midrule
      \textbf{Average} & \textbf{0.285} & \textbf{0.635} & \textbf{0.751} & \textbf{0.509} & \textbf{0.073} \\
      \textbf{Std}     & \textbf{0.333} & \textbf{0.015} & \textbf{0.175} & \textbf{0.164} & \textbf{0.015} \\
      \bottomrule
    \end{tabular}
    \caption{In-the-Wild Task Performance}
  \end{subtable}

  \begin{subtable}[t]{0.75\linewidth}
    \centering
    \vspace{5mm}
 
    \begin{tabular}{lcc}
      \toprule
      \textbf{Task} & \textbf{Robot} & \textbf{1:2} \\
      \midrule
      Spray                    & 0.120 & 0.520 \\
      Toy Cleanup              & 0.500 & 0.625 \\
      Bimanual Florist         & 0.063 & 0.623 \\
      Bimanual Clothes Folding & 0.198 & 0.740 \\
      \midrule
      \textbf{Average} & \textbf{0.220} & \textbf{0.627} \\
      \textbf{Std}     & \textbf{0.195} & \textbf{0.090} \\
      \bottomrule
    \end{tabular}
    \caption{In-the-Wild Extreme Task Performance} 
  \end{subtable}

 \caption{Performance Across Cotrain Ratios for Varying Deployment Conditions. Ratio is Robot:Human}
 \label{tab:cotrain_grouped}
\end{table}

\begin{table}[H]
\begin{center}

\begin{tabular}{@{}p{4.2cm}p{3.2cm}@{}}
\toprule
\textbf{Hyperparameter} & \textbf{Value} \\
\midrule
\multicolumn{2}{c}{\textbf{Training Configuration}} \\
\midrule
Optimizer & AdamW \\
Base Learning Rate & 3e-4 \\
Optimizer Momentum & $\beta_1, \beta_2 = 0.95, 0.999$ \\
Learning Rate Schedule & Cosine (diffusers) \\
Warmup Steps & 2000 \\
Total Steps & 70000 \\
Batch Size & 256 \\
Environment Frequency & 30 Hz \\
\midrule
\multicolumn{2}{c}{\textbf{Observation Settings}} \\
\midrule
Proprioception Horizon & 
\makecell[l]{1 (Spray, Toy, Pour) \\ 3 (Florist, Clothes)} \\
Image Horizon & 1 (all tasks) \\
Observation Resolution & 224×224 \\
Observation Dim &
\makecell[l]{9 (Spray, Toy, Pour) \\ 27 (Florist, Clothes)} \\
Action Dimension & 
\makecell[l]{26 (Spray, Toy, Pour) \\ 52 (Florist, Clothes)} \\
Action Chunk Size & 48  \\
\midrule
\multicolumn{2}{c}{\textbf{Action Chunking Transformer}} \\
\midrule
\# Encoder Layers & 4 \\
\# Decoder Layers & 6 \\
\# MHSA Heads & 8 \\
Feed-Forward Dim & 3200 \\
Hidden Dim (Token Dim) & 768 \\
Dropout & 0.1 \\
Feature Norm & LayerNorm \\
\midrule
\multicolumn{2}{c}{\textbf{Diffusion U-Net Policy}} \\
\midrule
Train Diffusion Steps & 100 \\
Eval Diffusion Steps & 16 \\
Down Channels & [256, 512, 1024] \\
Kernel Size & 3 \\
Groups (GN) & 8 \\
Dropout & 0.1 \\
Feature Norm & None \\
\bottomrule
\end{tabular}
\caption{\small Full training and architecture settings used across our experiments.}
\label{tab:full-hyperparams}
\end{center}
\vspace{-0.2in}
\end{table}

\end{document}